\documentclass{article}

\usepackage{PRIMEarxiv}

\usepackage[utf8]{inputenc} 
\usepackage[T1]{fontenc}    
\usepackage{hyperref}       
\usepackage{url}            
\usepackage{booktabs}       
\usepackage{amsfonts}       
\usepackage{nicefrac}       
\usepackage{microtype}      
\usepackage{lipsum}
\usepackage{fancyhdr}       
\usepackage{graphicx}       
\graphicspath{{media/}}     
\newcommand*{\thead}[1]{\multicolumn{1}{b{1.9cm}}{\centering\bfseries #1}}

\title{Iof-maint - Modular maintenance ontology
\thanks{\textit{\underline{Citation}}: 
\textbf{M.Hodkiewicz, C.Woods, M.Selway, M.Stumptner. 2024. IOF-Maint-Modular Maintenance Ontology. DOI:10.26182/chzp-vs60.}} 
}

\author{
  Melinda Hodkiewicz, Caitlin Woods \\
  The University of Western Australia \\
  Perth, Western Australia \\
  \texttt{\{caitlin.woods, melinda.hodkiewicz\}@uwa.edu.au} \\
   \And
  Matt Selway,  Markus Stumptner\\
  University of South Australia \\
  Adelaide, South Australia\\
  \texttt{\{matt.selway, markus.stumptner\}@unisa.edu.au} \\
}

\usepackage{array}
\usepackage{booktabs}
\usepackage{graphicx}
\usepackage{tikz}
\usepackage{comment}
\def\checkmark{\tikz\fill[scale=0.3](0,.35) -- (.25,0) -- (1,.7) -- (.25,.15) -- cycle;} 

\begin{document}
\maketitle

\begin{abstract}
In this paper we present a publicly-available maintenance ontology (Iof-maint). Iof-maint is a modular ontology aligned with the Industrial Ontology Foundry Core (IOF Core) and contains 20 classes and 2 relations. It provides a set of maintenance-specific terms used in a wide variety of practical data-driven use cases. Iof-maint supports OWL DL reasoning, is documented, and is actively maintained on GitHub. In this paper, we describe the evolution of the Iof-maint reference ontology based on the extraction of common concepts identified in a number of application ontologies working with industry maintenance work order, procedure and failure mode data. 
\end{abstract}


\section{Introduction}

The competitiveness of manufacturing and engineering organizations depend, in part, on how they maintain their assets to ensure availability to meet production, quality, cost and safety goals. Maintenance is defined as ``the actions intended to retain an item in, or restore it to, a state in which it can perform a required function''~\cite{ASIEC60300.3.14}. Almost all assets of any value need maintenance and, as a result, there has evolved organizational structures and processes to manage maintenance with theory and practice to support its delivery. Maintenance and asset management practices are documented in Standards~\cite{ASIEC60300.3.14,ISO55001} and by professional societies~\cite{SMRP,GFMAM_Maintenance}.

Data relevant to maintenance decision-making is located in sources such as computerised maintenance management systems (CMMS), requirements documentation, original equipment manufacturer manuals, process control systems, failure modes and effects analysis (FMEA) and failure investigations, to name just a few. However, finding this data and then being able to use with confidence due to data quality concerns continues to present challenges~\cite{hodkiewicz2014data,gitzel2015data,yang2020novel}. 

Most engineering organizations will have a CMMS whose data schemes conform to a common understanding of work flows in maintenance practice. Similarly the use of standards for processes such as FMEA impose a standard set of terms prescribed in, for example, IEC 60812:2018~\cite{IEC60812}. As a result, the data capture process for artifacts relating to maintenance work is remarkably consistent. The details of the work vary with asset type but the need to model concepts and workflow processes are common. This makes maintenance an ideal candidate for an ontology that can have wide-spread application as the classes used to describe the maintenance management process and information are commonly used by the maintenance community~\cite{karray2019romain}.

Engineering projects require multi-disciplinary teams, inside and across organisations, to exchange information at every stage of the asset life cycle, and there has long been interest in ontologies for the maintenance phase of the system~\cite{Ebrahimipour2010,matsokis2010ontology,karray2012formal,Kiritsis2013,ebrahimipour2016ontology,sanfilippo2019formal}. 
Moreover, operations and maintenance is often the most expensive and certainly the longest phase of an asset's life cycle during which a vast quantity of data is collected. As such, the potential business value of applying ontologies to enable semantic interoperability for industrial maintenance data and applications is very high.

This paper presents a reference ontology (Iof-maint) for asset maintenance, aligned to the Industrial Ontologies Foundry's (IOF) Core ontology. The IOF is responsible for a set of open reference ontologies to support the  needs of the manufacturing and engineering industries and advance data interoperability.

\section{Background}

\subsection{Background of Maintenance Management}

\begin{figure}
    \centering
    \includegraphics[width=1.\textwidth]{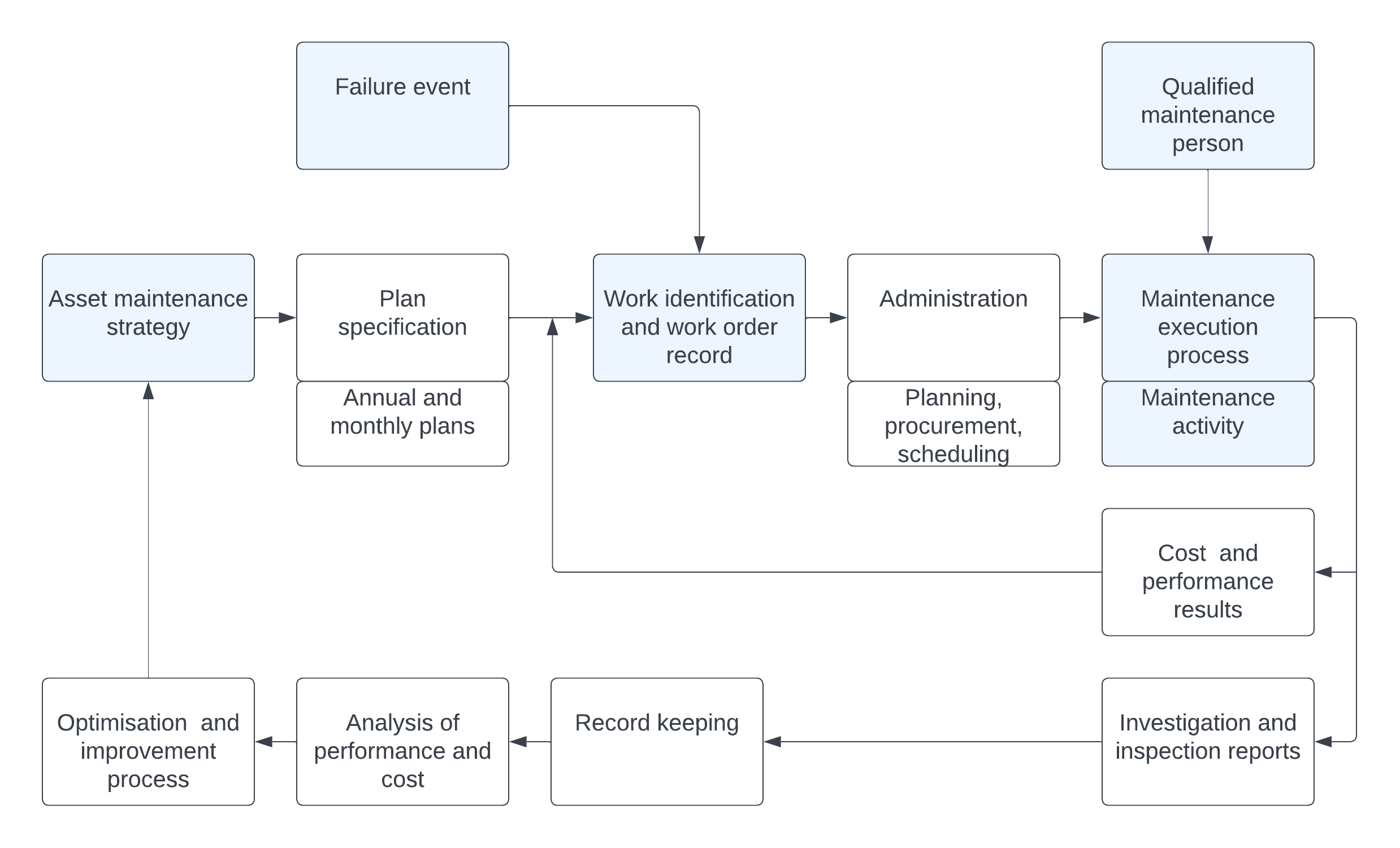}
    \caption{The maintenance management process, adapted from Coetzee~\cite{coetzee2004maintenance} and Karray \cite{karray2019romain}. The shaded boxes show the stages that have modelling features specific to maintenance management e.g.failure event and maintenance work orders, while the unshaded areas identify generic business process steps.}
    \label{fig:maint-manag}
\end{figure}

Maintenance management is a core business function in all asset-owning industries performed to ensure the asset delivers on its required product, service or function safely, cost effectively and meeting environmental and other regulations. The maintenance management process contains core elements as shown in Figure~\ref{fig:maint-manag}. Typical elements in this process include the development of a maintenance strategy for each asset, the outputs of strategy are combined into asset, system and production plans which inform scheduled activities such as inspections, servicing, repairs and replacements. These scheduled activities are initiated through the maintenance work order system which also captures required corrective orders resulting from failure events. All maintenance groups have planning, procurement and scheduling functions to enable maintenance to be executed in a timely, safe and coordinated fashion. Cost and performance reports along with reports generated in the maintenance process inform business functions such as analysis and improvement. The maintenance management process contains elements that are generic to the business function such as planning, scheduling, cost and performance analysis, these are shown as unshaded boxes in Figure~\ref{fig:maint-manag}. The shaded boxes represent elements that have characteristics in the way that they are done that are specific to maintenance and need special consideration in the modelling process. We have drawn on a number of maintenance practitioner texts and international standards to describe these concepts in the ontology, for example~\cite{CEN13306,GFMAM_Maintenance,palmer1999maintenance,ASIEC60300.3.14}.

Each asset has a primary \textit{function}, this describes the main reason(s) for owning or using the asset. It may also have a set of secondary functions due to the need to fulfill regulatory requirements and requirements concerning issues of protection, control, containment, comfort, appearance, energy efficiency, and structural integrity~\cite{SAEJA1012}. The \textit{maintenance action} performed on an asset depends on the \textit{function} to be maintained and the consequence of a functional failure. Since the 1970s, the most common approach to developing an asset's \textit{maintenance strategy} is a process called Reliability-centered Maintenance (RCM). In RCM, the emphasis is on selecting an appropriate strategy to manage the risk of each consequential failure mode. The output of the RCM process is the identification of \textit{maintenance strategies}, \textit{maintenance activities} and timings for each \textit{maintainable item}. The concepts described above for \textit{function, maintenance strategy, maintenance activity, and maintainable item} are specific to maintenance and need to be represented in a maintenance domain-specific ontology.

Preventative, inspection, and failure-finding activities identified in the strategies are triggered at the appropriate time in the CMMS as a \textit{maintenance work order} record (MWO). Corrective work is triggered by a \textit{failure event} on a \textit{maintainable item}. Once work is identified, either through failure events or actions triggered by maintenance strategies, then the work is planned and scheduled. Planning involves the procurement of parts and the organization of personnel, tools, other resources and the availability of the asset. Once complete work can be scheduled. Finally, the value-adding step is when the work is executed by maintenance personnel. After execution work there should be analysis of the quality and effectiveness of the maintenance work and opportunities for improvement identified. We suggest that the processes of planning, scheduling, and execution are generic to many industrial processes. As a result they do not need to be developed specifically for this maintenance ontology. Instead existing classes in other modular domain-independent ontologies that are aligned to the same top-level ontology can be re-used to represent these processes.

\subsection{Ontologies for Maintenance}

The ROMAIN maintenance reference ontology published in 2019 was the first reference ontology for maintenance management \cite{karray2019romain} to make .owl files publicly available\footnote{\url{https://github.com/HediKarray/ReferenceOntologyOfMaintenance}}. ROMAIN used a hybrid approach, based on a top-down alignment to an open source top-level ontology, the Basic Formal Ontology (BFO), and a bottom-up focus on classes grounded in maintenance practice. The authors constrained the scope of the ontology to the classes that are unique to the maintenance management practice, as described in the section above, but in order to link to data for a use case demonstration, a number of additional (case-specific) classes need to be included.

Prior to the release of ROMAIN, maintenance had been focus in asset/ product life cycle ontologies taking an asset/product-centric view rather than a maintenance management view \cite{matsokis2010ontology,karray2012formal,Kiritsis2013,otte2019ontological,may2022semantic}. These works do not provide publicly available RDF implementations of the ontology for industry practitioners to test.

ROMAIN, while influential academically, focused on the conceptual model with limited validation on application use cases. Subsequent work to develop use cases, validate, and refine the ontology occurred in the context of collaboration with IOF development. It is the results of this 3 year industry-academic collaboration, to produce a shared, open and professionally managed  maintenance reference ontology aligned to BFO and interoperable with other industry ontologies, that are presented below.   

\subsection{The Industrial Ontologies Foundry}

\begin{figure}[ht]
    \centering
    \includegraphics[width=0.8\textwidth]{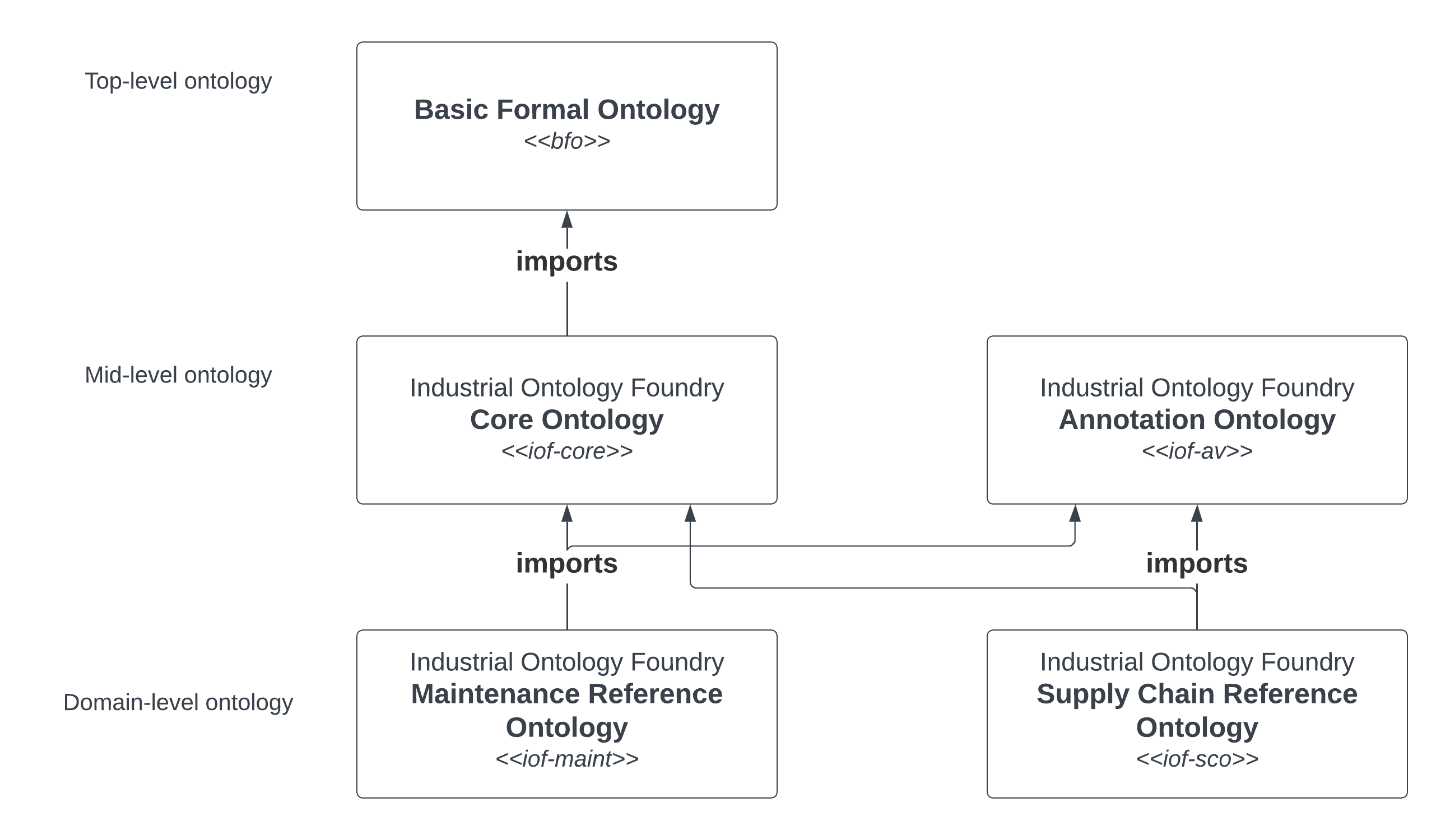}
    \caption{IOF Ontologies import structure}
    \label{fig:bfo_overview}
\end{figure}

\begin{figure}[!h]
    \centering
    \includegraphics[width=0.8\textwidth]{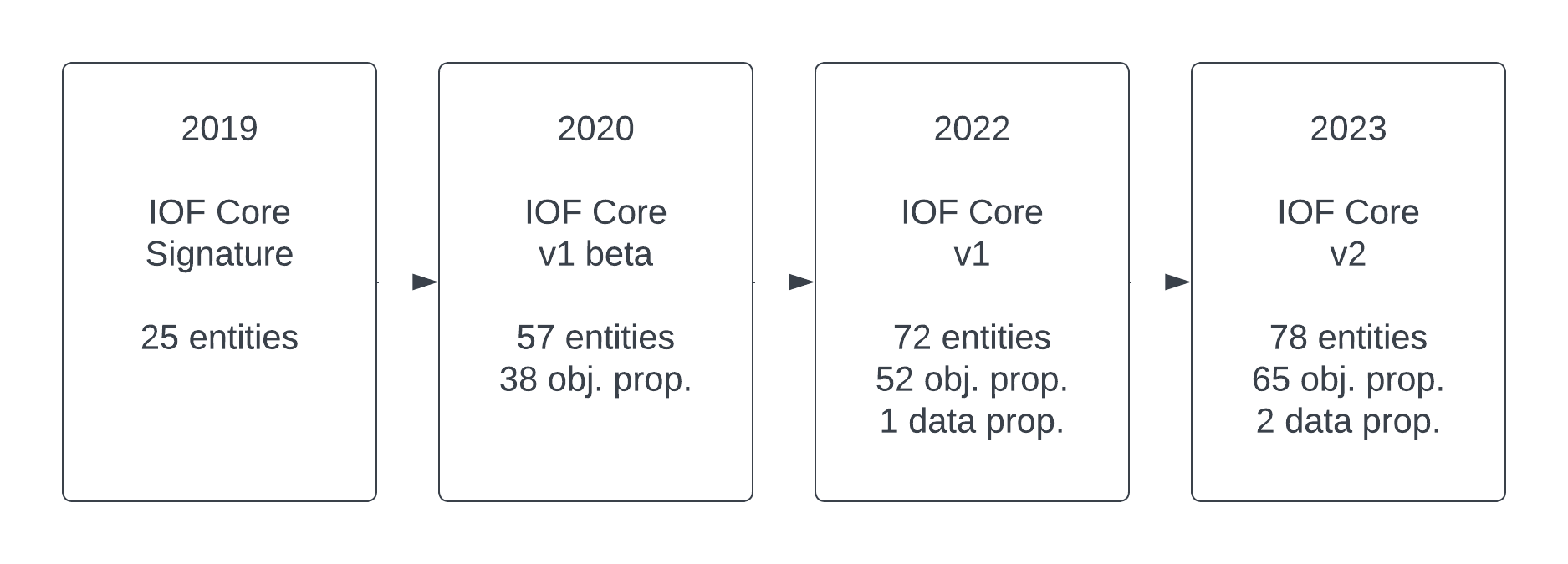}
    \caption{Evolution of the Industrial Ontology Core ontology}
    \label{fig:iof_core_evolution}
\end{figure}

The IOF initiative was formed to create a set of mutually consistent, open-source ontologies to provide a basis for data and knowledge representation across the manufacturing and engineering domains~\cite{drobn2022fomi}. The IOF's goal is to create a suite of principles-based ontologies aligned with a domain-independent top-level ontology called the Basic Formal Ontology~\cite{otte2022bfo}. The hub contains reference ontologies, currently the Core, Annotation, Supply Chain and Maintenance Ontologies, as shown in Figure~\ref{fig:bfo_overview}. These ontologies are non-redundant in the sense that they assert no terms in common. All the ontologies in the suite are interoperable, in a precisely defined sense, due to their adherence to a common set of principles. In the IOF community multiple parties agree to use one another’s ontologies, to share a common set of principles, and to share the work of revising both ontologies and principles as these are tested in use with ever-expanding bodies of data, thus ensuring that the result achieves a required degree of interoperability. For example, the evolution of the IOF Core ontology, illustrated in Figure~\ref{fig:iof_core_evolution}, occurred as parties identified the needs of their domains and came together to refine and agree on the common elements. Each ontology in the foundry is managed by a working group responsible for development, documentation, use case development and maintenance. Each working group is composed of domain experts and ontology practitioners. There is a technical board for peer review and quality control. 

In its early days the IOF considered a number of top-level ontologies as candidates to align their work to and in 2018 the community adopted BFO. BFO uses a realism-based approach in terms of the existence in time of classes and represents things as they are in reality~\cite{Arp2015}.  BFO is a specification of one particular conceptualization of the world that represents reality as two disjoint categories of continuant (independent and dependent continuants, attributes, and locations) and occurrent (processes and temporal regions). BFO-2020 has 35 terms and 45 relations and is defined in both OWL and first-order logic. BFO provides a high level abstraction and a set of constraints which, when adhered to, allow one BFO-aligned ontology to work with another BFO-aligned ontology. BFO is documented in an International Standard~\cite{IEC21838-2} which is important for industrial adoption, especially in regulated industries.





\subsection{Language, Namespaces, URIs and Class Axiom Expressions}

BFO and IOF are implemented using Web Ontology Language (OWL) version 2 Description Logic (a.k.a.\ OWL 2 DL). OWL 2 DL adds expressivity to the associated Resource Description Framework (RDF) and RDF Schema (RDFS). OWL, RDF and RDFS allows one to define hierarchies of classes and relationships, and individuals, link individuals to data values, assign values to XML Schema Definition datatypes, assert relationships between classes, and between object properties, or between classes and object properties, as well as create useful annotations for classes, individuals, and relationships~\cite{rudnicki2019overview}. 

IOF makes use of terms from the following schemas:
\begin{itemize}
    \item bfo:	\url{http://purl.obolibrary.org/obo/},
    \item owl: \url{http://www.w3.org/2002/07/owl},
    \item rdf: \url{http://www.w3.org/1999/02/22-rdf-syntax-ns},
    \item rdfs:	\url{http://www.w3.org/2000/01/rdf-schema},
    \item skos:	\url{http://www.w3.org/2004/02/skos/core},
    \item xml:	\url{http://www.w3.org/XML/1998/namespace}
    \item xsd:	\url{http://www.w3.org/2001/XMLSchema}, and 
    \item  dcterms:	\url{http://purl.org/dc/terms/}.
\end{itemize}
 	 	
Following the Semantic Web rules for linked data (Berners-Lee, 2006; Bizer, et al., 2009), the IOF adopts HTTP-based URIs as unambiguous identifiers for entities. IOF utilizes a common namespace under which all IOF ontologies reside\footnote{https://spec.industrialontologies.org/iof/ontology/}. The path component for an ontology identifier then includes the working group (e.g., `core', `maintenance', etc.) followed by the ontology name. The ontologies are given convenient namespace prefixes that should be used consistently, such as `iof-core' for the IOF-Core ontology.
The identifier of a class, property, or individual is introduced following the forward slash (/) at the end of the ontology identifier. For example, the IOF-Core class Maintainable Material Item has this URI \footnote{https://spec.industrialontologies.org/iof/ontology/core/Core/MaintainableMaterialItem}.

\section{Approach}

The iof-maint ontology has evolved from a set of four use cases documented in peer-reviewed published works~\cite{woods2023activity,woods2023procedure,woods2021notion,hodkiewicz2021ontology,karray2019romain}. Each use case is described by an application ontology and supported by industrial data produced in the maintenance work management process. Each use case aligns to an upper ontology. 

Classes that appear multiple times in these application ontologies or are deemed central to the maintenance management process are lifted to the iof-maint ontology as a common resource. We also move object properties to iof-maint where they are essential in defining these classes (and do not already exist in BFO or IOF-Core).

\begin{figure}
    \centering
    \includegraphics[width=1.\textwidth]{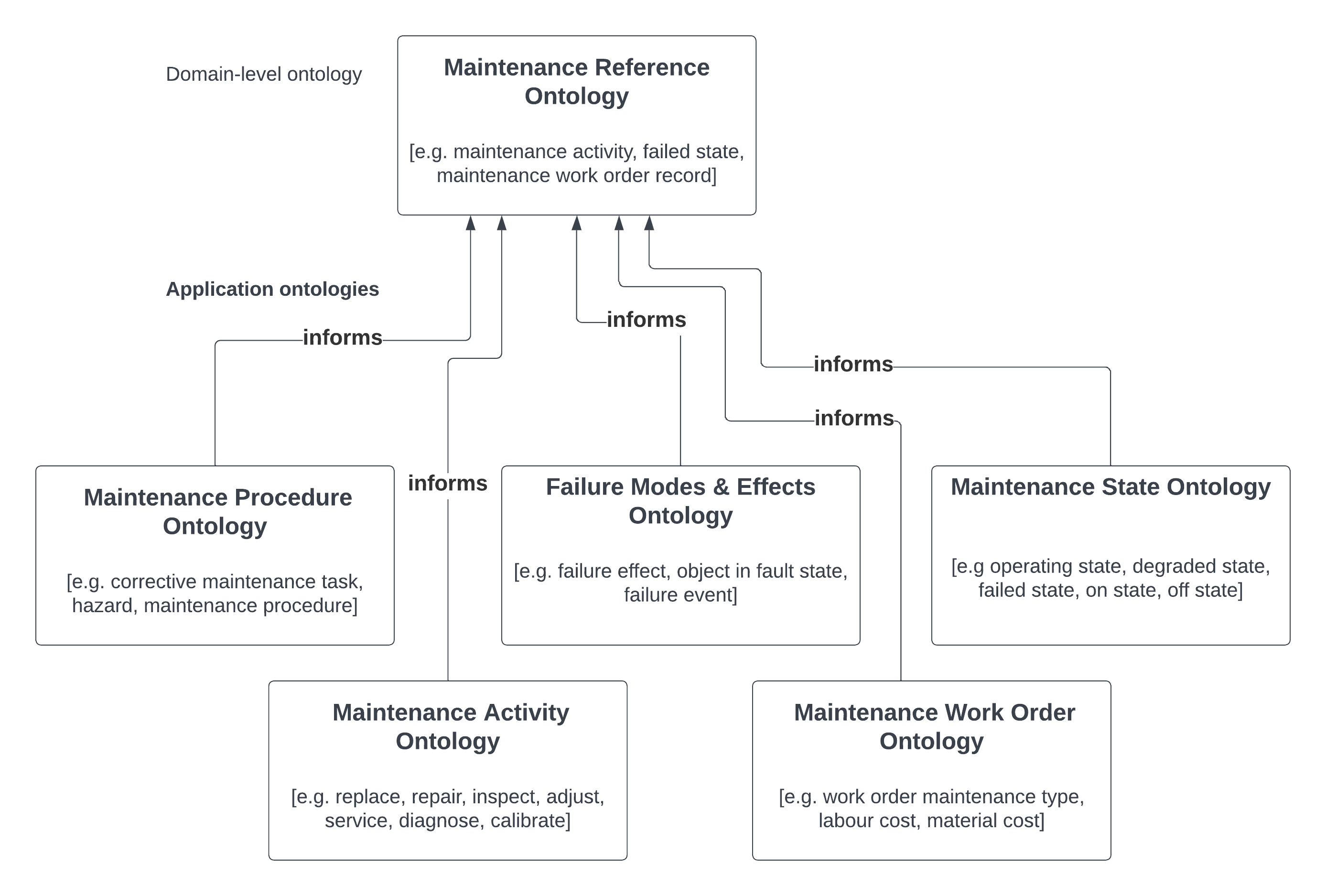}
    \caption{The application ontologies developed in the ontology engineering process for iof-maint}
    \label{fig:alignment}
\end{figure}

Figure~\ref{fig:alignment} shows the published application ontologies on which iof-maint is based. The \textit{Maintenance Activity Ontology}~\cite{woods2023activity}, \textit{Maintenance Work Order Ontology}~\cite{woods2023activity} and \textit{Maintenance State}~\cite{woods2021notion} ontologies model knowledge found in maintenance work order documents. The \textit{Maintenance Procedure Ontology}~\cite{woods2023procedure} models maintenance procedure documentation used by maintenance technicians. Finally, the \textit{Failure Modes and Effects Ontology}~\cite{hodkiewicz2021ontology} models FMEA documentation that maintenance engineers produce to as an input to maintenance strategy development. Table~\ref{tab:class-map} shows how each of the terms in iof-maint map to each of the domain ontologies.
Note that the \textit{Maintenance strategy} class is not mapped to any of the application ontologies: this is because it was identified as a key maintenance related class but is yet to be leveraged in a use case, although definitions do refer to it in expectation of its use in future work.

\begin{table}[!ht]
\centering
\caption{Classes in iof-maint that are moved from application ontologies}
\label{tab:class-map}
\small
\begin{tabular}{>{\raggedright\arraybackslash}p{4.1cm}p{1.5cm}p{1.5cm}p{1.5cm}p{1.5cm}p{1.4cm}}
\thead{Class} & \thead{Maint.\ Procedure} & \thead{Maint.\ Activity} & \thead{Maint.\ State} & \thead{FMEA} & \thead{Maint.\ Work Order} \\ \toprule
Failure effect & & & & \centering\checkmark & \\ \midrule
Failure process & \centering\checkmark & & \centering\checkmark & & \\ \midrule
Functioning process & & & \centering\checkmark & & \\ \midrule
Undesirable disposition & & & \centering\checkmark & \centering\checkmark & \\ \midrule
Disposition to exhibit undesirable behaviour & & & \centering\checkmark  & & \\ \midrule
Disposition to fail & & & \centering\checkmark  & & \\ \midrule
Required function & & & \centering\checkmark  & \centering\checkmark  & \\ \midrule
Failure event & \centering\checkmark  & & \centering\checkmark  & \centering\checkmark  & \\ \midrule
Maintenance strategy & & & & & \\ \midrule
Failure mode code & \centering\checkmark  & & & \centering\checkmark & \\ \midrule
MWO record & & & & & \checkmark \\ \midrule
Maintenance state & & & \centering\checkmark & & \\ \midrule
Operating maintenance state & & & \centering\checkmark & & \\ \midrule
Degraded maintenance state & & & \centering\checkmark & & \\ \midrule
Failed maintenance state & & & \centering\checkmark & \centering\checkmark & \\ \midrule
Maintenance activity & & \centering\checkmark & & & \\ \midrule
Maintenance process & \centering\checkmark & & & & \\ \midrule
Qualified maintenance person & \centering\checkmark & & & & \\ \midrule
Qualification specification & \centering\checkmark & & & & \\ \midrule
Supporting maintenance activity & & \centering\checkmark & & & \\ \bottomrule                                 
\end{tabular}
\end{table}

In Sections \ref{sec:uc-1}, \ref{sec:uc-2},  \ref{sec:uc-3} and  \ref{sec:uc-4} we describe each of the use cases that we used to develop the five domain ontologies pictured in Figure~\ref{fig:alignment}.

\subsection{Maintenance activity and end-of-life events} \label{sec:uc-1}
This Maintenance Activity Ontology~\cite{woods2023activity} uses existing international engineering standards \cite{ISO15926-4} \cite{ISO14224} and a set of +800,000 maintenance work orders to develop and define a set of maintenance activity terms (and synonyms for these). This paper also includes a Maintenance Work Order Ontology (capturing structured information such as the labour and material costs of a work order) to check that the maintenance activities documented in the work order correctly represent the work performed on the equipment. The value of this work is for a) checking data quality for maintenance work orders and b) repeatably identifying end-of-life events for equipment, information used in maintenance strategy optimisation.

\subsection{Digitisation of maintenance procedures} \label{sec:uc-2}

Maintenance procedure documentation contain valuable information, such as required resources and safety information for executing maintenance work. These documents, however, are paper-based in many organisations. This maintenance procedure ontology~\cite{woods2023procedure} answers competency questions that support engineers, schedulers, and technicians in their typical uses of procedure documentation at work. This ontology is aligned to the ISO TC 15926-14~\cite{kluwer2008iso} industrial ontology which is conceptually similar to BFO.

\subsection{Maintenance state} \label{sec:uc-3} 

An important part of maintenance is understanding the conditions that need to be met for equipment to perform its required function, and how changes in state affect an equipment item's ability to perform this function. This paper discusses the notion of a `Maintenance State' and presents a BFO-aligned ontology (the Maintenance State Ontology~\cite{woods2021notion}) for understanding the effects of state changes for maintenance equipment. 

\subsection{Failure modes and effects analysis} \label{sec:uc-4}
FMEA documentation is generated by maintenance engineers to understand possible equipment failures (and the effects of these failures). Maintenance engineers design and use this documentation to develop maintenance strategies. The FMEA Ontology~\cite{hodkiewicz2021ontology} captures and reasons over the knowledge found in FMEA documentation. It is aligned to the ISO TC 15926-14~\cite{kluwer2008iso} industrial ontology.

\section{Maintenance Reference Ontology}

The Maintenance Reference Ontology (iof-maint) supports modelling of concepts associated with asset reliability and maintenance management. It is relevant for data captured at the design stage of the asset life cycle (e.g. in failure modes and effects analysis) and in the operations and maintenance phase of the life cycle (e.g. asset reliability and the maintenance activities associated with delivering value from use of the assets). This Maintenance Reference Ontology is a minimal ontology aligned to the IOF Core ontology (iof-core) and by extension BFO. It deliberately captures only concepts frequently occurring in the maintenance phase of the asset life cycle, regardless of whether the data originates in the maintenance phase. This minimal ontology supports the use of modular ontologies for specific classes of interest and maintenance-application ontologies as described in the previous section. 

\subsection{Maintenance-relevant Ontology terms}

There are 20 classes in the Maintenance Reference Ontology (iof-maint) as shown in Table \ref{tab:maint_class_summary}. In addition there are a number of equipment and maintenance related terms in the IOF Core ontology. Examples of these terms are in Table \ref{tab:iof_class_summary}. These terms sit in the Core module as they are necessary to model equipment and maintenance related terms and data in a wide variety of (non-maintenance) applications, for example supply chain and logistics and production planning.  The two terms lifted from an earlier version of iof-maint to the current version of iof-core are \textit{maintainable material item} and \textit{maintainable material item role}. One can see the value of the modular approach in that terms such as \textit{material component}, \textit{requirement specification}, \textit{physical location identifier} and \textit{plan specification} sit in the Iof-core and can be accessed by all compliant ontologies.

\begin{table}[!ht]
\centering
\caption{Maintenance Reference Ontology terms in iof-maint }
\label{tab:maint_class_summary}
  \begin{tabular}{p{6cm}p{6cm}}
    \toprule
$(1)$ Failure effect  &	$(2)$ Failure process \\
$(3)$ Functioning process & $(4)$ Undesirable disposition \\
$(5)$ Disposition to exhibit undesirable \newline $\-$ $\-$ $\-$ $\-$ $\-$ behaviour &	$(6)$ Disposition to fail   \\
$(7)$ Required function	& $(8)$ Failure event\\	
$(9)$ Maintenance strategy & $(10)$ Failure mode code\\	
$(11)$ Maintenance work order record &	$(12)$ Qualified maintenance person \\
$(13)$ Operating maintenance state & 	$(14)$ Degraded maintenance state \\
$(15)$ Failed maintenance state & $(16)$ Maintenance state \\
$(17)$  Maintenance process  & 	$(18)$  Qualification specification \\
$(19)$  Maintenance activity   & $(20)$ Supporting maintenance activity  \\
 \bottomrule
\end{tabular}
\end{table}

\begin{table}[!ht]
\centering
\caption{Equipment, plan, requirement and other maintenance-related terms in IOF Core (iof-core)}
\label{tab:iof_class_summary}
  \begin{tabular}{p{6cm}p{6cm}}
    \toprule
$[1]$ Maintainable material item     &	$[2]$ Material artefact \\
$[3]$ Material component  & $[4]$ Maintainable material item role \\
$[5]$ Equipment role &	$[6]$ Objective specification \\
$[7]$ Plan specification   & $[8]$ Requirement specification \\
$[9]$ Qualification specification & $[10]$ Organisation identifier \\
$[11]$ Physical location identifier &	$[14]$ Measured value expression \\
\bottomrule
\end{tabular}
\end{table}

\section{Formal Definitions and Axioms} 

Details of 4 of the 20 classes in the Maintenance Reference Ontology are provided in Table \ref{tab:definitions}. Each defined class has natural language, first order logic and a semi-formal definition and a code to identify if it is primitive or not. Each primitive has a natural language definition and necessary or sufficient axioms as appropriate. For space reasons the remaining 16 terms plus the FOL and axioms for each  term are not included here in the text but are readily accessible on GitHub accessible via \footnote{\url{https://spec.industrialontologies.org/iof/ontology/maintenance/MaintenanceReferenceOntology/}}. 

\begin{table}[ht!]
\centering
\label{tab:definitions}
\caption{Examples of 4 terms from the iof-maint ontology selected to illustrate the nature of maintenance-specific concepts. Full definitions of the remaining terms including first-order logic are available on GitHub}
\begin{tabular}{p{3cm}p{12.5cm}}
\toprule
\multicolumn{2}{l}{\textbf{Failure effect}}   \\ \midrule
NL Def.         &   process that is the consequence of failure, within or beyond the boundary of the failed item \vspace{0.2cm}    \\  
Semi-formal Def.      &     if x is a `failure effect' then x is a `process' that is 'preceded by' some 'failure event' or `failure process'                                          \vspace{0.2cm}     \\
Primitive  & true \\
Examples & leaking pipe, erratic operation, equipment does not run \\ 
\midrule
\multicolumn{2}{l}{\textbf{Failure process}}   \\ \midrule
NL Def.         &   process that changes some quality of an item causing the the item to become degraded or failed \vspace{0.2cm}    \\  
Semi-formal Def.      &     every instance of `failure process' is defined as exactly an instance of `process' that `realizes' some `disposition to fail'                                          \vspace{0.2cm}     \\
Primitive  & false \\
Synonym & functional failure \\
Examples & short circuiting process, deformation process, corrosion process \\ 
\midrule
\multicolumn{2}{l}{\textbf{Functioning process}}   \\ 
\midrule
NL Def.         &   process that enables an item to perform a function \vspace{0.2cm}    \\  
Semi-formal Def.      &     every instance of `functioning process' is defined as exactly an instance of `process' that `realizes' some `function'              \vspace{0.2cm}     \\
Primitive  & false \\
Examples & explosion, seizure, loss of power, loss of control \\ 
\midrule
\multicolumn{2}{l}{\textbf{Maintenance state}}   \\ \midrule
NL Def.         &   stasis that holds during a temporal interval when the realizable functions and capabilities of the participating item, or the grade of realization of those functions and capabilities, remain unchanged \vspace{0.2cm}    \\ 
Semi-formal Def.      &    every instance of `maintenance state' is defined as exactly an instance of `operating state' or exactly an instance of `degraded state' or exactly an instance of `failed state'        \vspace{0.2cm}     \\
Primitive  & false \\
Examples & is broken in two, is running at desired speed \\  
\bottomrule
\end{tabular}
\label{tbl:2}
\end{table}

\section{Discussion}

This reference ontology meets the specifications set out by~\cite{kendall2019ontology} that the ontology should be: 
\begin{itemize}
    \item encoded formally in a declarative knowledge representation language,
    \item syntactically well-formed for the language, as verified by an appropriate syntax checker or parser,
    \item logically consistent, as verified by a language-appropriate theorem or reason prover,
    \item and meet business or application requirements as demonstrated through testing.
\end{itemize}

The governance processes established by the IOF ensure that each ontology must meet the requirements set out above. This conformance is assessed by the IOF Technical Operating Board and a confirmed by a vote by all members. One realisation from this work is that the last bullet point in the above list is challenging, for this reference ontology as a stand-alone artifact, without addition terms being defined. We have demonstrated real business-relevant questions can be answered but only when this reference ontology is combined with application ontologies such as was done for the maintenance procedure, failure modes and effects analysis and maintenance activity ontologies described earlier. These application ontologies contain the additional classes,  instance data and SWRL rules necessary to address specific business questions. The IOF-core industrial ontology, this iof-maint maintenance reference ontology and iof-scro a supply chain reference modules are part of a rapidly evolving landscape for industrial reference ontology development to support industrial applications.

In addition to the iof-maint reference ontology being professionally maintained and made publicly available on GitHub, the ontology is also available on the Ontocommons Portal - a website produced with EU funding to support and accelerate the uptake of ontology in industry \footnote{http://industryportal.enit.fr/ontologies/}. The Ontocommons EU-funded project conducted an assessment (available on the Ontocommons weblink in the footnote) of the iof-maint ontology against FAIR principles based on 13 Findable questions, 13 Accessible questions, 15 Interoperable questions, and 20 Reusable questions \cite{amdouni2022faire}. The Iof-maint ontology FAIR score is 271. This compares favourably with other well known ontologies such as the Semantic Sensor Network Ontology SSN (FAIR score of 256) and QUDT of (FAIR score of 187).

The availability and suitability of reference ontologies is an increasingly important value proposition for organisations seeking to build application ontologies. The import and reuse of domain terms means that only a small number of additional terms may be required to model the use case. For example if the maintenance activity \cite{woods2023activity} was to be built today (rather than earlier, before iof-maint) only 15 classes (describing various types of maintenance activity) would have to be defined. This building-block approach, re-using existing (upper-ontology aligned) modules makes ontology creation more efficient. The high quality documentation provided on the iof-maint web site is also an aid for users. 

Of interest for iof-maint is the ongoing work of the Industrial Data Ontology (IDO) community. IDO has emerged from the oil and gas sector and is an evolution of work previously documented in the ISO 15926-14 Technical Report \cite{kluwer2008iso} and in application use cases~\cite{brynildsen2023iswc,skjaeveland2021ottr,hodkiewicz2021ontology,cameron2022digital}. The emergence of IOF and IDO, both industry supported, provides choice for modellers. Which one they choose depends on a number of factors. The BFO 2020 version adopted by IOF has time-based relations, whereas IDO does not. BFO has a large suite of BFO-aligned ontologies, especially in the science areas, whereas IDO has concentrated on material master data and models relevant to high hazard  industries. While the IDO ontology is currently freely available through the PoscCaesar website \footnote{https://rds.posccaesar.org/ontology/lis14/} the recent acceptance of IDO into the International Standards Organisation's standardisation pathway as ISO CD 23726-3 means that open access to future versions of both the IDO ontology and its documentation will be limited as ISO operates only paid access to its standards. Regardless, both IOF and IDO have important markets and the authors of iof-maint are involved in both groups. Hence current work is focused on producing an IDO-aligned variant of this maintenance reference ontology ensuring that industry teams modelling maintenance data can continue to use the ontology regardless of the upper ontology alignment selected.

\section{Conclusion}

 For the design engineering, process, infrastructure and manufacturing sectors the business value of being able to locate, assess, combine and reason over equipment-related maintenance data held in their data lakes and data stores is considerable. Industry interest is high in the potential for explicit conceptual representations, in the form of named entity-based knowledge graphs and ontologies, to assist with 1) semantic interoperability of relational database, unstructured text and multi-modal data held in enterprise data stores, and 2) supporting Large Language Models (LLM) training, tuning, and deployment. 

 This iof-maint modular maintenance reference ontology  enables efficient modelling of maintenance related data. Iof-maint is encoded formally in OWL 2 DL to enable reasoning, it is publicly available, maintained by a professional organisation, modular to work with other BFO aligned ontologies, and has been demonstrated to meet business and application requirements. Our goal is that the Iof-maint resource will provide a common semantic base for organisations building enterprise knowledge graphs that draw data from computerised maintenance management systems thus enabling engineers to retrieve and manipulate information with confidence in its semantic intent. 

\section*{Acknowledgments}
The authors would like to thank the members of the Technical Oversight Board of the Industrial Ontologies Foundry for their support of this work, their discussions and reviews have added immeasurably to the outcome. 

\bibliographystyle{unsrt}  
\bibliography{references}  

\end{document}